\documentclass[letterpaper, 10 pt, conference]{ieeeconf}  

\IEEEoverridecommandlockouts                              

\usepackage{graphicx} 
\usepackage{epsfig} 
\usepackage{mathptmx} 
\usepackage{times} 
\usepackage{amsmath} 
\usepackage{amssymb}  
\usepackage{hyperref}
\usepackage[caption = false]{subfig}
\usepackage[noadjust]{cite}

\newcommand{\etal}{\textit{et~al}.}

\title{\LARGE \bf
Robotic Grasp Detection using Deep Convolutional Neural Networks
}

\author{Sulabh Kumra$^1$ and Christopher Kanan$^2$
\thanks{$^{1}$Sulabh Kumra is with Xerox Corporation, Webster, NY, USA and the Department of Electrical Engineering, Rochester Institute of Technology, Rochester, NY, USA.
        {\tt\small sk2881@rit.edu}}%
        \thanks{$^{2}$Christopher Kanan is with Carlson Center for Imaging Science at the Rochester Institute of Technology, Rochester, NY, USA.
        {\tt\small kanan@rit.edu}}%
}

\begin{document}

\maketitle
\thispagestyle{empty}
\pagestyle{empty}

\begin{abstract}

Deep learning has significantly advanced computer vision and natural language processing. While there have been some successes in robotics using deep learning, it has not been widely adopted. In this paper, we present a novel robotic grasp detection system that predicts the best grasping pose of a parallel-plate robotic gripper for novel objects using the RGB-D image of the scene. The proposed model uses a  deep convolutional neural network to extract features from the scene and then uses a shallow convolutional neural network to predict the grasp configuration for the object of interest. Our multi-modal model achieved an accuracy of 89.21\% on the standard Cornell Grasp Dataset and runs at real-time speeds. This redefines the state-of-the-art for robotic grasp detection.

\end{abstract}

\section{INTRODUCTION}

Robotic grasping lags far behind human performance and is an unsolved problem in the field of robotics. When humans see novel objects, they instinctively know how to grasp to pick them up. A lot of work has been done related to robotic grasping and manipulation  ~\cite{saxena2008robotic,jiang2011efficient,Ciocarlie2014,lenz2015deep,Redmon}, but the problem of real-time grasp detection and planning is still a challenge. Even the current state-of-the-art grasp detection techniques fail to detect a potential grasp in real-time. The robotic grasping problem can be divided into three sequential phases: grasp detection, trajectory planning, and execution. Grasp detection is  a visual recognition problem in which the robot uses its sensors to detect graspable objects in its environment. The sensors used for perceiving the robot's environment are typically 3-D vision systems or RGB-D cameras. The key task is to predict potential grasps from sensor information and map the pixel values to real world coordinates. This is a critical step in performing a grasp as the subsequent steps are dependent on the coordinates calculated in this step. The calculated real world coordinates are then transformed to position and orientation for the robot's end-of-arm tooling (EOAT). An optimal trajectory for the robotic arm is then planned to reach the target grasp position. Subsequently, the planned trajectory for the robotic arm is executed using either an open-loop or a closed loop controller. In contrast to an open-loop controller, a closed-loop controller receives continuous feedback from the vision system during the entire grasping task. The additional processing needed to handle the feedback is computationally expensive and can drastically affect the speed of the task. 

\begin{figure}
\begin{center}
\includegraphics[width=0.72\linewidth]{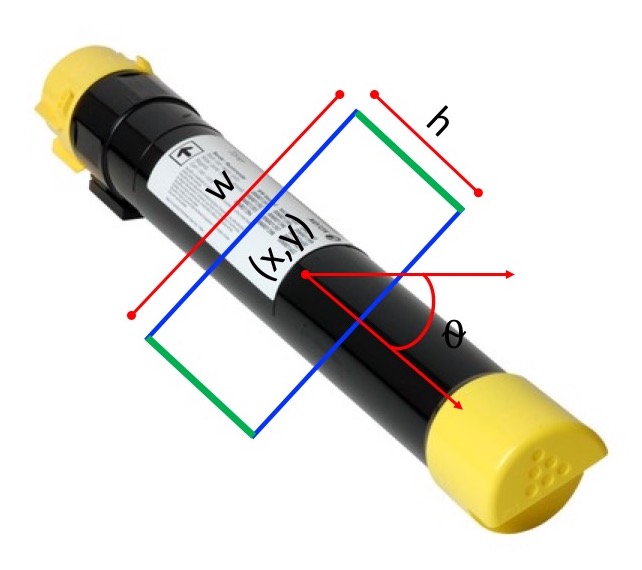}
\end{center}
   \caption{An example grasp rectangle for a potential good grasp of a toner cartridge. This is a five-dimensional grasp representation, where green lines represent parallel plates of gripper, blue lines correspond to the distance between parallel plates of the grippers before grasp is performed, (x,y) are the coordinates corresponding to the center of grasp rectangle and $\theta$ is the orientation of the grasp rectangle with respect to the horizontal axis.}
\label{fig:rit_mug}
\end{figure}

In this paper, we target the problem of detecting a `good grasp' from RGB-D imagery of a scene. Fig. \ref{fig:rit_mug} shows a five-dimensional grasp representation for a potential good grasp of a toner cartridge. This five-dimensional representation gives the position and orientation of a parallel plate gripper before the grasp is executed on an object. Although, it is a simplification of the seven-dimensional grasp representation introduced by Jiang \etal ~\cite{jiang2011efficient}, Lenz \etal ~showed that a good five-dimensional grasp representation can be projected back to a seven-dimensional grasp representation that can be used by a robot to perform a grasp~\cite{lenz2015deep}. In addition to low computational cost, this reduction in dimension allows us to detect grasps using RGB-D images. In this work, we use this five-dimensional grasp representation for predicting the grasp pose.

\begin{figure}
\vspace{2mm}
\begin{center}
\includegraphics[width=1\linewidth]{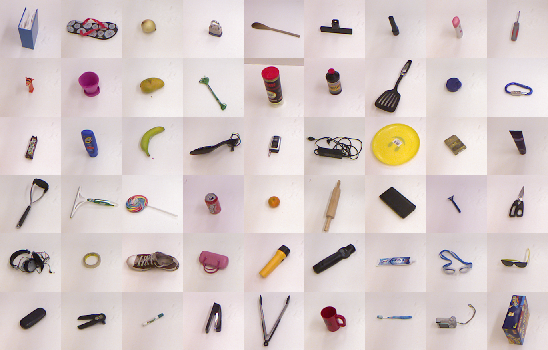}
\end{center}
   \caption{Sample images from the Cornell Grasp Dataset.}
\label{fig:dataset}
\end{figure}

We introduce a novel approach for detecting good robotic grasps for parallel plate grippers using the five-dimensional representation. Our approach uses two 50-layer deep convolutional residual neural networks running in parallel to extract features from RGB-D images, with one network analyzing the RGB component and the other analyzing the depth channel. The outputs of these networks are then merged, and fed into another convolutional network that predicts the grasp configuration. We compare this approach to others in the literature, as well as a uni-modal variation of our model that uses only the RGB component. Our experiments are done on the standard Cornell Grasp Dataset. Example images from the dataset are shown in Fig.~\ref{fig:dataset}. Our experiments show that the proposed architecture outperforms the current state-of-the-art methods in terms of both accuracy and speed.

\section{BACKGROUND}

Deep learning~\cite{cite-key} has significantly advanced progress on multiple problems in computer vision~\cite{NIPS2012_4824,russakovsky2015imagenet,Taigman_2014_CVPR} and natural language processing~\cite{sutskever2014sequence,cho2014learning,socher2011dynamic}. These results have inspired many robotics researchers to explore the applications of deep learning to solve some of the challenging problems in robotics. For example, robot localization is moving from using hand-engineered features~\cite{johns2014generative} to deep learning features~\cite{sunderhauf2015performance}, deep reinforcement learning is being used for end-to-end training for robotic arm control~\cite{levine2016end}, multi-view object recognition has achieved state-of-the-art performance by deep learning camera control~\cite{johns2016pairwise}, reinforcement learning has been used to learn dual-arm manipulation tasks~\cite{kumra2015dual}, and  autonomous driving has been tackled by using deep learning to estimate the affordances for driving~\cite{chen2015deepdriving}.

A major challenge with deep learning is that it needs a very large volume of training data. However, large datasets with manually labeled images are unavailable for most robotics applications. In computer vision, transfer learning techniques are used to pre-train deep convolutional neural networks on some large dataset, e.g., ImageNet, which contains 1.2 million images with 1000 categories~\cite{imagenet_cvpr09}, before the network is trained on the target dataset~\cite{yosinski2014transferable}. These pre-trained models are either used as an initialization or as a fixed feature extractor for the task of interest.

The most common approach for 2-D robotic grasp prediction is a sliding window detection framework. In this framework, a classifier is used to predict whether a small patch of the input image has a good potential grasp for an object. The classifier is applied to a number of patches on the image and the patches that get high scores are considered as good potential grasps. Lenz \etal~used this technique with convolutional neural networks as a classifier and got an accuracy of 75\%~\cite{lenz2015deep}. A major drawback of their work is that it runs at a speed of 13.5 seconds per frame, which is is extremely slow for a robot to find where to move its EOAT in real-time applications. In \cite{Redmon}, this method was accelerated by passing the entire image through the network at once, rather than passing several patches.

A significant amount of work has been done using 3-D simulations to find good grasps \cite{bohg2010learning,krainin2011autonomous,5509508,5649406}. These techniques are powerful, but most of them rely on a known 3-D model of the target object to calculate an appropriate grasp. However, general purpose robots should be able to grasp unfamiliar objects without object's 3-D model. Jincheng \etal~showed that deep learning has the potential for 3-D object recognition and pose estimation, but their experiments only used five objects and their algorithm is computationally expensive~\cite{yu2013vision}. Recent work by Mahler \etal~uses a cloud-based robotics approach to significantly reduce the number of samples required for robust grasp planning~\cite{7487342}. Johns \etal~generated their training data by using a physics simulation and depth image simulation with 3-D object meshes to learn grasp score which is more robust to gripper pose uncertainty~\cite{johns2016deep}.

Grasp point detection technique proposed by Jeremy \etal ~\cite{maitin2010cloth} has very high precision of 92\%, but it only works with cloth towels and cannot be used as a general purpose grasp detection technique. Another grasp pose detection technique was introduced by Gualtieri \etal ~\cite{gualtieri2016high} for removing objects from a dense cluster. The technique was evaluated only on a small set of objects using a research robot.

We take a different approach, instead of using AlexNet for feature extraction, as used in \cite{lenz2015deep}, \cite{Redmon} and \cite{7487517}, we use the current state-of-the-art deep convolutional neural network known as ResNet \cite{he2015deep}. We also introduce a multi-modal model which extracts features from both RGB and Depth images to predict the grasp configuration.


\section{PROBLEM FORMULATION}
The robotic grasp detection problem can be formulated as finding a successful grasp configuration $g$ for a given image $I$ of an object. A five-dimensional grasp configuration $g$ is represented as:
\begin{equation}
g = f(x, y, h, w, \theta)
\end{equation}
where $(x,y)$ corresponds to the center of grasp rectangle, $h$ is the height of parallel plates, $w$ is the maximum distance between parallel plates and $\theta$ is the orientation of grasp rectangle with respect to the horizontal axis. $h$ and $w$ are usually fixed for a specific robot EOAT. An example of this representation is shown in Fig.~\ref{fig:rit_mug}. 

We focus on planer grasps as Lenz \etal~showed that a five-dimensional grasp configuration can be projected back to a seven-dimensional configuration for execution on a real robot. To solve this grasp detection problem, we take a different approach, explained in section \ref{approach}.


\section{APPROACH}
\label{approach}
Deep convolutional neural networks (DCNNs) have outperformed the previous state-of-the-art techniques to solve detection and classifications problems in computer vision. In this paper, we use DCNNs to detect the target object from the image and predict a good grasp configuration. We propose a single step prediction technique instead of the two step approach used in \cite{lenz2015deep} and \cite{Redmon}.
These methods ran a simple classifier many times on small patches of the input image, but this is slow and computationally expensive. Instead, we feed the entire image directly into DCNN to make grasp prediction on complete RGB-D image of the object. This solution is simpler and has less overhead.

Theoretically, a DCNN should have better performance with increased depth because it provides increased representational capacity. However, our current optimization method, stochastic gradient decent (SGD) is not an ideal optimizer. In experiments, researchers found that increased depth brought increased training error, which is not in-line with the theory~\cite{he2015deep}. The increased training error indicates that the ultra-deep network is very hard to optimize. This means that identity map is very hard to obtain in a convolutional neural network by end-to-end training using SGD. Therefore, we use residual layers as in ResNet~\cite{he2015deep}, which reformulates the mapping function between layers, using the function given by eq.\eqref{eq:resnet}.

Similar to previous works, we assume that the input image contains only one graspable object and a single grasp has to be predicted for the object. The advantage of this assumption is that we can look at the complete image and make a global grasp prediction. This assumption may not be possible outside the experimental conditions and we would have to come up with a model that has to first divide the image into regions, so each region contains only one object.

\begin{figure}
\vspace{2mm}
\begin{center}
\includegraphics[width=0.8\linewidth]{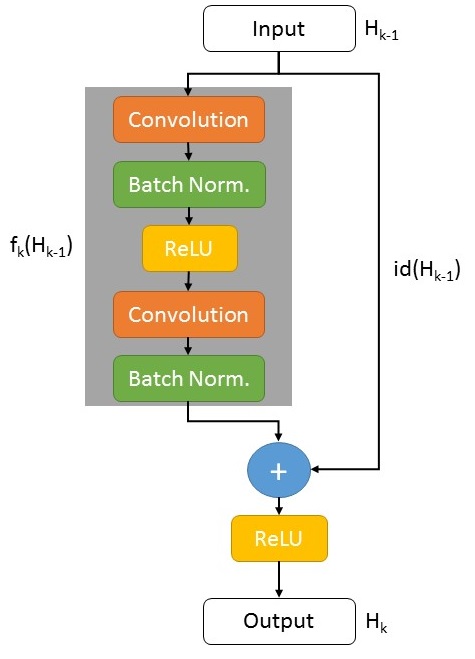}
\end{center}
   \caption{Example residual block in ResNet.}
\label{fig:resnet}
\end{figure}

\subsection{Architecture}
Our model is much deeper as compared to the previous approaches (e.g., \cite{lenz2015deep, 7487517, Redmon}). Instead of using an eight layer AlexNet, we use ResNet-50, a fifty layer deep residual model, to solve this grasp detection problem. The ResNet architecture uses the simple concept of residual learning to overcome the challenge of learning an identity mapping. A standard feed-forward CNN is modified to incorporate skip connections that  bypass a few layers at a time. Each of these skip connections gives rise to a residual block, and the convolution layers predict a residual that is added to the block's input.  The key idea is to bypass the convolution layers and the non-linear activation layers in $k^{th}$ residual block, and let through only the identity of the input feature in the skip connection. Fig.~\ref{fig:resnet} shows an example of a residual block with skip connections. The residual block is defined as:
\begin{equation}
\label{eq:resnet}
H_k = F(H_{k-1}, {W_k}) + H_{k-1}
\end{equation}
where, $H_{k-1}$ is the input to the residual block, $H_k$ is the output of the block, and $W_k$ are the weights learned for the mapping of function $F$. We encourage the readers to see ~\cite{he2015deep} for more details on the ResNet architecture.

We introduce two different architectures for robotic grasp prediction: uni-modal grasp predictor and multi-modal grasp predictor. The uni-modal grasp predictor is a 2D grasp predictor that uses only single modality (e.g., RGB) information from the input image to predict the grasp configuration, where as the multi-modal grasp predictor is a 3-D Grasp Predictor that uses multi-modal (e.g., RGB and Depth) information. In the next two subsections, we discuss these two architectures in detail.

\begin{figure}
\vspace{2mm}
\begin{center}
\includegraphics[width=0.7\linewidth]{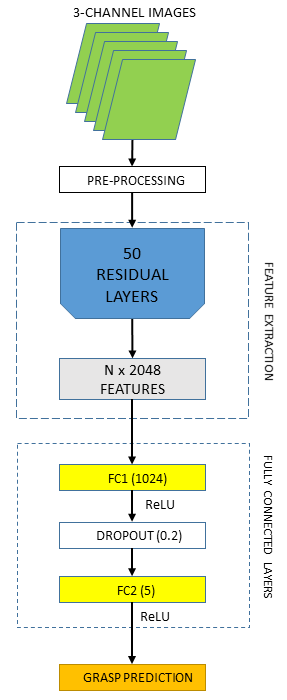}
\end{center}
   \caption{Complete architecture of our uni-modal grasp predictor.}
\label{fig:2dmodel}
\end{figure}

\subsection{Uni-modal Grasp Predictor}
Large-scale image classification datasets have only RGB images. Therefore, we can pre-train our deep convolutional neural networks with 3-channels only. We introduce a uni-modal grasp predictor model which is designed to detect grasp using only three channels (RGB or RGD) of the raw image. Fig. \ref{fig:2dmodel} shows the complete architecture of our uni-modal grasp predictor. A ResNet-50 model that is pre-trained on ImageNet is used to extract features from the RGB channels of the image. For a baseline model, we use a linear SVM as classifier to predict the grasp configuration for the object using the features extracted from the last hidden layer of ResNet-50. In our uni-modal grasp predictor, the last fully connected layer of ResNet-50 is replaced by two fully connected layers with  rectified linear unit (ReLU) as activation functions. A dropout layer is also added after the first fully connected layer to reduce over-fitting. We use SGD to optimize our training loss and mean squared error (MSE) as our loss function.

The 3-channel image is fed to the uni-modal grasp predictor, which uses the residual convolutional layers to extract features from the input image. Last fully connected layer is the output layer, which predicts the grasp configuration for the object in the image. During training time, weights of convolutional layers in ResNet-50 are kept fixed and only the weights of last two fully connected layers are tuned. The weights of the last two layers are initialized using Xavier weight initialization.

\subsection{Multi-modal Grasp Predictor}
\begin{figure}[!ht]
\vspace{2mm}
\begin{center}
\includegraphics[width=0.8\linewidth]{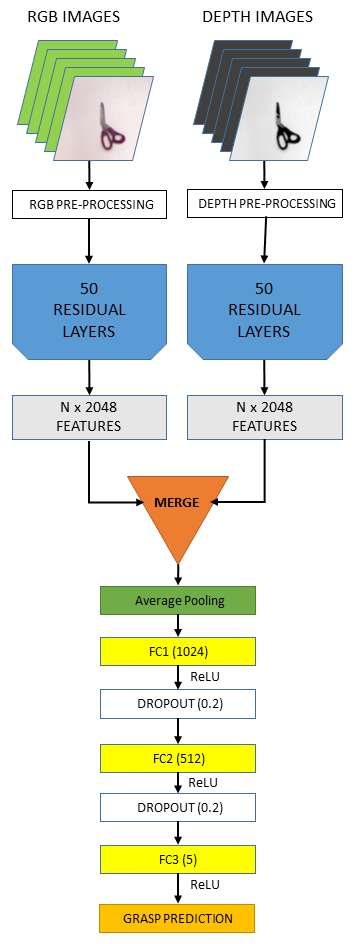}
\end{center}
   \caption{Complete architecture of our multi-modal grasp predictor.}
\label{fig:3dmodel}
\end{figure}

We also introduce a multi-modal grasp predictor, which is inspired by the RGB-D object recognition approach introduced by Schwarz \etal~\cite{schwarz2015rgb}. The multi-modal grasp predictor uses multi-modal (RGB-D) information from the raw images to predict the grasp configuration. The raw RGB-D images are converted into two images. The first is a simple RGB image and other is a depth image converted into a 3-channel image. This depth to 3-channel conversion is done similar to a gray to RGB conversion. These two 3-channel images are then given as input to two independent pre-trained ResNet-50 models. The ResNet-50 layers work as feature extractors for both the images. Similar to the uni-modal grasp predictor, features are extracted from the second last layer of both the ResNet-50 networks. The extracted features are then normalized using L2-normalization. The normalized features are concatenated together and feed into a shallow convolutional neural network with three fully connected layers. The fully connected layers use ReLU activation functions. We added a dropout layer after first and second fully connected layers of the shallow network to reduce over-fitting. Similar to the uni-modal model, we used SGD as the optimizer and MSE as the loss function. Fig.~\ref{fig:3dmodel} shows the complete architecture of our multi-modal grasp predictor.

By using two DCNNs in parallel, the model was able to extract features from both RGB and depth images. Therefore, enabling the model to learn multimodal features from the RGB-D dataset. Weights of the two DCNNs are initialized using the pre-trained ResNet-50 models and the weights of the shallow network are initialized using Xavier weight initialization. The weights are fine tuned during training.

As a simple baseline, we also applied a linear SVM classifier to the L2-normalized RGB DCNN and depth DCNN features to predict the grasp configuration for the object in the image.

\section{EXPERIMENTS}
\subsection{Dataset}
\begin{figure}
\vspace{2mm}
\begin{center}
\includegraphics[width=0.8\linewidth]{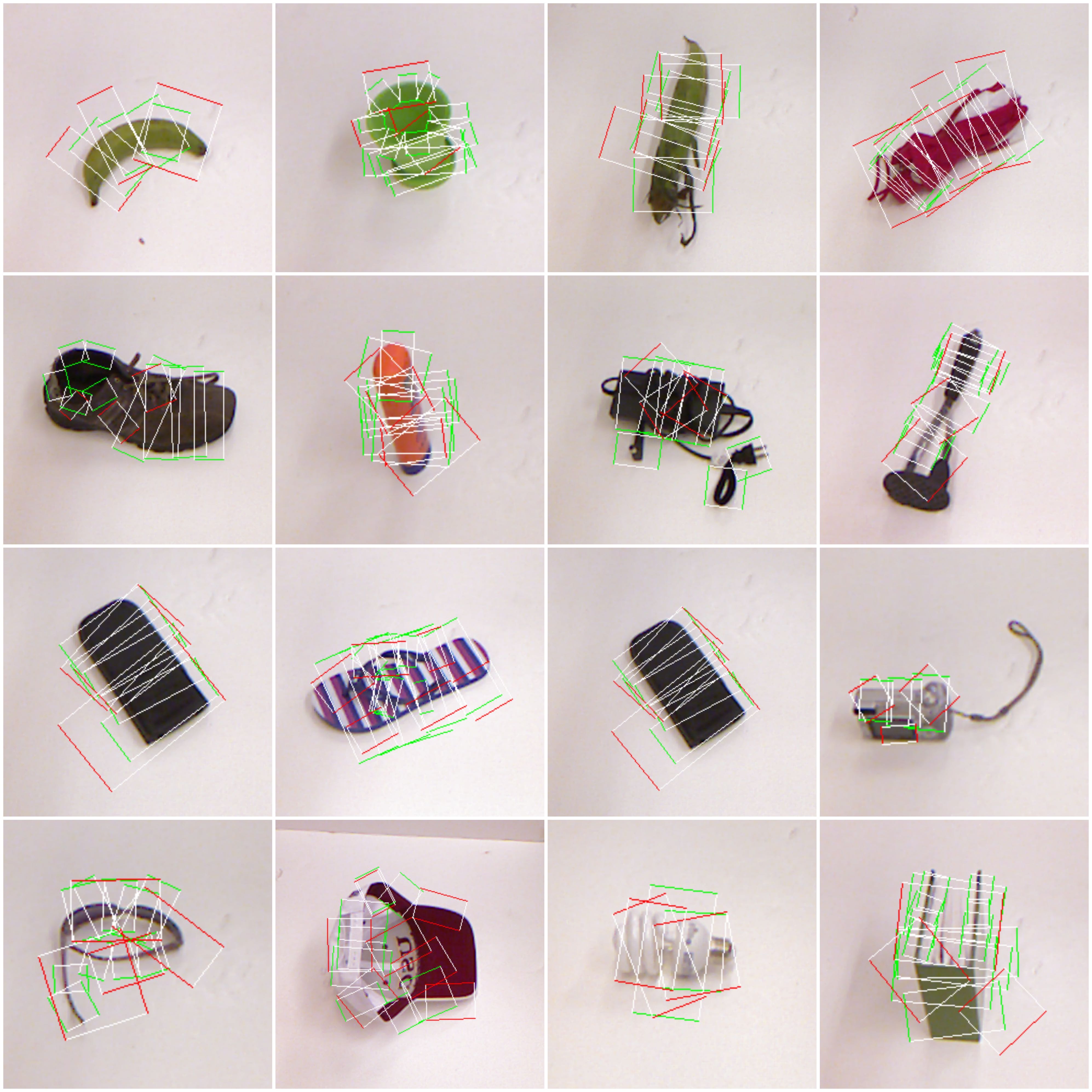}
\end{center}
   \caption{Ground truth grasps using the rectangular metric for a subset of the Cornell grasp dataset.}
\label{fig:GraspCollage}
\end{figure}

For comparing our method with others, we test our architecture on the standard Cornell Grasp Dataset. The dataset is available at \url{http://pr.cs.cornell.edu/grasping/rect_data/data.php}. This dataset consists of 885 images of 240 different objects. Each image has multiple grasp rectangles labeled as successful (positive) or failed (negative), specifically selected for parallel plate grippers. In total, there are 8019 labeled grasps with 5110 positive and 2909 negative grasps. Fig. \ref{fig:GraspCollage} shows the ground truth grasps using the rectangular metric for this dataset.

Similar to previous works, we have used five-fold cross validation for all our experiments. The dataset is split in two different ways:  
\begin{enumerate}
\item \textbf{Image-wise split} \\
Image-wise splitting splits all the images in the dataset randomly into the five folds. This is helpful to test how well did the network generalize to the objects it has seen before in a different position and orientation.
\item \textbf{Object-wise split} \\
Object-wise splitting splits all the object instances randomly and all images of an object are put in one validation set. This is helpful to test how well did the network generalize to objects it has not seen before.
\end{enumerate}

\subsection{Data Pre-processing}
We perform a minimal amount of data pre-processing before feeding it into the DCNN. The input to our DCNN is a patch around the grasp point, extracted from a training image. The patch is re-sized to $224 \times 224$, which is the input image size of the  ResNet-50 model. The depth image is re-scaled to range 0 to 255. There are some pixels in depth image that have a NaN value as they were occluded in the original stereo image. These pixels with NaN value were replaced by zeros.

\subsection{Pre-training}
Pre-training is necessary when the domain-specific data available is limited as in the Cornell grasp dataset. Therefore, ResNet-50 is first trained on ImageNet. We assume that most of the filters learned are not specific to the ImageNet dataset and only the layers near the top exhibit specificity for classifying 1000 categories. The DCNN will learn universal visual features by learning millions of parameters during this pre-training process. We then grab the features from the last layer and feed it to our shallow convolutional neural network. It is important to note that the ImageNet dataset has only RGB images and thus the DCNN will learn RGB features only.

\subsection{Training}
For training and validation of our models we used Keras deep learning library, which is written in Python and running on top of Theano. Experiments were performed on a CUDA enabled NVIDIA GeForce GTX 645 GPU with Intel(R) Core(TM) i7-4770 CPU @ 3.40GHz. Although, GPUs are currently not an integral part of robotic systems, they are getting popular in vision-based robotic systems because of the increased computational power.

Our training process was divided into two stages, in the first stage, only the shallow network is trained, and in the second stage the complete network is trained end-to-end. To train our uni-modal grasp predictor, we used SGD to optimize the model with hyper parameters in first stage set as: learning rate = 0.001, decay = 1e-6, momentum = 0.9, mini-batch size = 32 and maximum number of epoch = 30. For the multi-modal grasp predictor, we used the following hyper parameters in first stage: learning rate = 0.0006, decay = 1e-6, momentum = 0.9, mini-batch size = 32 and maximum number of epoch = 50. For fine-tuning the network in the second phase, we use a much lower learning rate and plateau the learning rate if the training loss does not decreases.

\begin{table*}
\vspace{2mm}
\begin{center}
\caption{Grasp Prediction Accuracy on the Cornell Grasp Dataset}
\label{tab:results}
\begin{tabular}{|l|l|c|c|}
\hline
\textbf{Authors} & \textbf{Algorithm} & \multicolumn{2}{|c|}{\textbf{Accuracy (\%)}} \\
\cline{3-4}
 & & Image-wise split & Object-wise split \\
\hline
& Chance & 6.7 & 6.7 \\
Jiang \etal \cite{jiang2011efficient} & Fast Search & 60.5 & 58.3\\
Lenz \etal \cite{lenz2015deep} & SAE, struct. reg. two-stage & 73.9 & 75.6 \\
Redmon \etal \cite{Redmon} & AlexNet, MultiGrasp & 88.0 & 87.1 \\
Wang \etal \cite{wang2016robot} & Two-stage closed-loop, with penalty & 85.3 & - \\
Asif \etal \cite{asif2017rgb} & STEM-CaRFs (Selective Grasp) & 88.2 & 87.5 \\
\hline
 & \textbf{Uni-modal Grasp Predictor} & & \\
 & ResNet-50, SVM - RGB (\textit{Baseline}) & 84.76 & 84.47 \\
 & ResNet-50, ReLU, ReLU - RGB & 88.84 & 87.72 \\
Ours & ResNet-50, tanh, ReLU - RGD & 88.53 & 88.40 \\
 &  \textbf{Multi-Modal Grasp Predictor} &  & \\
 & ResNet-50x2, linear SVM - RGB-D & 86.44 & 84.47 \\
 & ResNet-50x2, ReLU, ReLU, ReLU - RGB-D & \textbf{89.21} & \textbf{88.96} \\
\hline
\end{tabular}
\end{center}
\end{table*}

\subsection{Evaluation}
Prior works have used two different performance metrics for evaluating grasps on the Cornell grasp dataset: rectangle metric and point metric. The point grasp metric compares the distance between the center point of predicted grasp and the center point of all the ground truth grasps. A threshold is used to consider the success of grasp, but past work did not disclose these threshold values. Moreover, this metric does not consider the grasp angle, which is an essential parameter for grasping. The rectangle grasp metric consider complete grasp rectangle for evaluation and a grasp is considered to be a good grasp if the difference between the predicted grasp angle and the ground truth grasp angle is less than $30^{\circ}$, and the Jaccard similarity coefficient of the predicted grasp and ground truth grasp is more than $25\%$. Jaccard similarity coefficient or the Jaccard index measures similarity between the predicted grasp and ground truth grasp, and is defined as:
\begin{equation}
J(\widehat{\theta}, \theta) = \frac{| \widehat{\theta} \cap \theta|}{| \widehat{\theta} \cup \theta|}
\end{equation}
where $ \widehat{\theta}$ is the predicted grasp and $\theta$ is the ground truth grasp. As the rectangle metric is better at discriminating between 'good' and 'bad' grasp, we use this metric for our experiments. For all of our models, we select the best scored grasp rectangle using the rectangle metric for predicting the grasp.


\section{RESULTS}
Table \ref{tab:results} shows a comparison of our results with the previous work for the rectangle metric grasp detection accuracy on the Cornell RGB-D grasp dataset. Across the board, both of our models outperform the current state-of-the-art robotic grasp detection algorithms in terms of accuracy and speed. Results for the previous work are their self-reported scores. Tests were performed with image-wise split and object-wise split to test how well the network can generalize to different grasp features.

We present results of various versions of uni-modal and multi-modal grasp predictors. This is done by changing the information fed to the input channels. The RGB version of uni-modal grasp predictor uses only RGB channels of the input image. In the RGD version, we replace the blue channel of the input image with the re-scaled depth information. The baseline model of uni-modal grasp predictor got an accuracy of 84.76\%. Our uni-modal grasp predictor with RGB data got an accuracy of 88.84\% and the same model with RGD data achieved an accuracy of 88.53\%. In contrast to prior work, replacing blue channel with depth did not help our model. This is mainly because the ResNet was trained with RGB images and the features learned in RGB are not same as the features extracted from RGD.

\begin{figure}
\begin{center}
\includegraphics[width=0.9\linewidth]{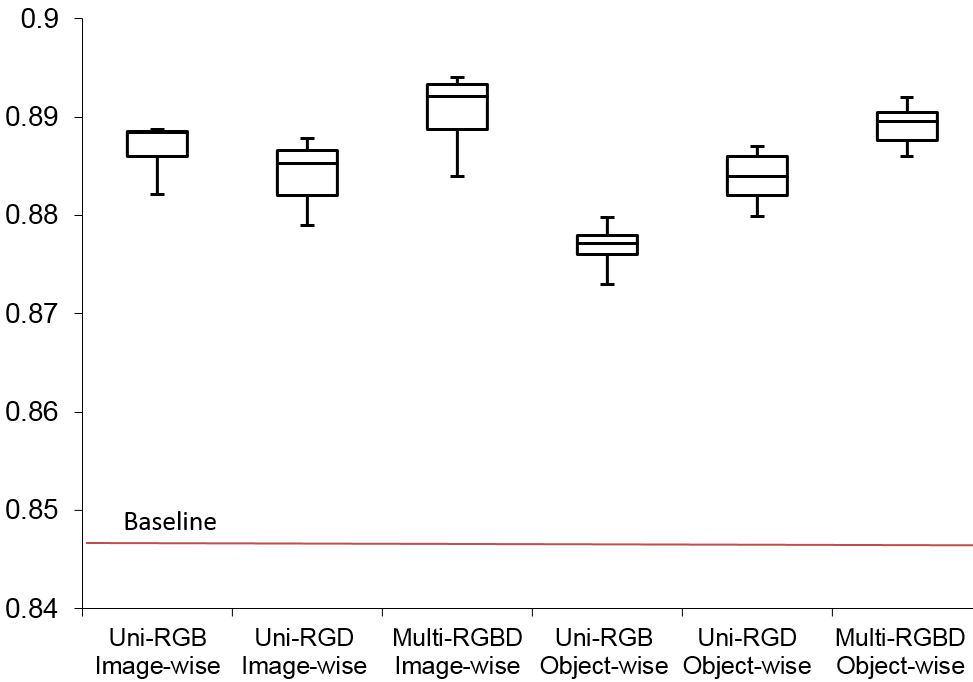}
\end{center}
\caption{Accuracy comparison of models.}
\label{fig:boxplot}
\end{figure}

The baseline multi-modal grasp predictor used RGB-D data and got an accuracy of 86.44\%, which sets a new baseline for performance in RGB-D robotic grasp detection. Our multi-modal grasp predictor achieved an accuracy of 89.21\%, which is the new state-of-the-art performance for RGB-D robotic grasp detection. We also tried replacing the ResNet-50 model with a pre-trained VGG16 model. Although, it performed better than previous models, it did not perform better than our multi-modal model. Fig. \ref{fig:boxplot} shows an accuracy comparison of all the proposed models in this paper using 5-fold cross validation. Overall, our multi-modal grasp predictor performed the best with the Cornell grasp dataset.

Table \ref{tab:speed} shows the grasp prediction speeds for our models and compares it with previous work. Both of our models are faster than previous methods. Our uni-modal grasp predictor runs 800 times faster than the two-stage SAE model by Lenz \etal. The main reason for this boost in speed is replacing the sliding window classifier based approach by a single pass model. We also used GPU hardware to accelerate computation and that can be another reason for faster computation.

\begin{table}
\vspace{2mm}
\begin{center}
\label{tab:speed}
\caption{Grasp Prediction Speed}
\begin{tabular}{|l|c|}
\hline
\textbf{Method} & \textbf{Speed (fps)} \\
\hline
Lenz \etal \cite{lenz2015deep} & 0.02 \\
Redmon \etal \cite{Redmon} & 3.31 \\
Wang \etal \cite{wang2016robot} & 7.10\\
\hline
Uni-modal Grasp Predictor & 16.03 \\
Multi-modal Grasp Predictor & 9.71 \\
\hline
\end{tabular}
\end{center}
\end{table}

\begin{figure}
\centering
\subfloat[True Positive]{\includegraphics[width = 1.2in]{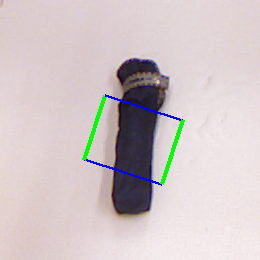} \label{predict_a}}
\subfloat[False Positive]{\includegraphics[width = 1.2in]{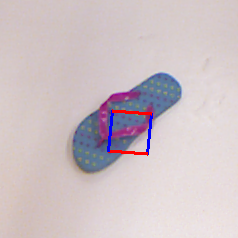} \label{predict_b}} \\
\subfloat[True Negative]{\includegraphics[width = 1.2in]{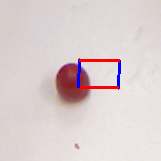} \label{predict_c}}
\subfloat[False Negative]{\includegraphics[width = 1.2in]{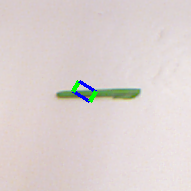} \label{predict_d}}
\caption{Examples of predicted graspability using the modified multi-modal grasp predictor.}
\label{fig:predict}
\end{figure}

We made a modification to the multi-modal model to predict the graspability, i.e. whether an object is graspable for a specific grasp rectangle or not. This was done by replacing the last fully connected layer by a dense layer with binary output and used softmax as the activation function. We were able to achieve an accuracy of 93.4\%, which is at par with the current state-of-the-art. Fig.~\ref{fig:predict} shows some examples of predicted graspability using the modified multi-modal grasp predictor. A green box means that a successful grasp was predicted and a red box means an unsuccessful grasp was predicted. The false negative (Fig.~\ref{predict_b}) and false negative (Fig.~\ref{predict_d}) are the incorrect predictions. In Fig.~\ref{predict_b} we believe that the model failed to understand the depth features of the slipper strap, using which the grippers can grasp the slipper. Whereas, in Fig.~\ref{predict_d} the model failed to understand the orientation $\theta$ of the grasp rectangle with respect to the object. Other than some tricky examples such as these, the model predicts the graspability of different types of objects with high accuracy.

\section{DISCUSSION}
We show that deep convolutional neural networks can be used to predict the graspability and grasp configuration for an object. Our network is 6 times deeper as compared to the previous work by Lenz \etal~ and we made an improvement of 14.94\% for image-wise split and 13.36\% for object-wise split. This shows that going deeper with network and using skip connections helps the model learn more features from the grasp dataset.

Our results show that high accuracy can be achieved with our multi-modal model and that it can be used to predict the graspability and grasp configuration for the objects that the model has not seen before. The uni-modal model got the best accuracy when used with RGB data and image-wise split of dataset. Whereas, the multi-modal model performed the best with RGB-D data and object-wise split of the dataset.

\begin{figure}
\centering
\subfloat[Bowl]{\includegraphics[width = 1.2in]{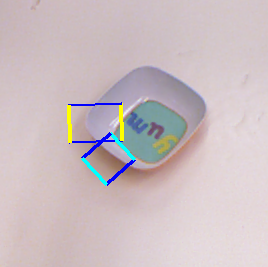} \label{predict_1}}
\subfloat[Headphones]{\includegraphics[width = 1.2in]{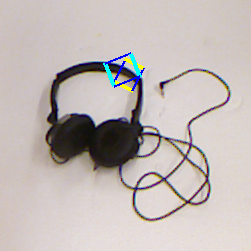} \label{predict_2}} \\
\subfloat[Tape]{\includegraphics[width = 1.2in]{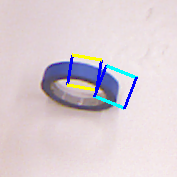} \label{predict_3}}
\subfloat[Lock]{\includegraphics[width = 1.2in]{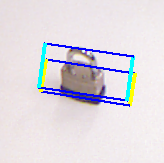} \label{predict_4}}
\caption{Uni-modal VS multi-modal grasp predictor.}
\label{fig:compare}
\end{figure}

Fig. \ref{fig:compare} shows cases when multi-modal grasp predictor (cyan and blue grasp rectangles) produces a viable grasp, while the uni-modal grasp predictor (yellow and blue grasp rectangles) fails to produce a viable grasp for the same object. In some of these cases, the grasp produced by the uni-modal predictor might be feasible for a robotic gripper, but the grasp rectangle produced by the multi-modal predictor represents a grasp which would clearly be successful.

Due to unavailability of a pre-trained ResNet model for depth data, both the ResNet-50 models used in the multi-modal model were pre-trained on ImageNet. This may not be the best model for the depth image as the model is only trained on RGB images and will not have depth specific features. In the future, we would like to pre-train the model on a large-scale RGB-D grasp dataset like \cite{levine2016learning} and then use it to predict RGB-D grasps. Moreover, if we have a large-scale RGB-D grasp dataset, we can modify our uni-modal model to take a four channel input and predict grasps using all four channels. In this case, the input size for the network will be ($224 \times 224 \times 4$) and we can pass RGB as first three channels and depth as the fourth channel.


\section{CONCLUSION}
In this paper, we presented a novel multi-modal robotic grasp detection system that predicts the graspability of novel objects for a parallel plate robotic gripper using RGB-D images, along with a uni-modal model that uses RGB data only. We showed that DCNNs can be used in parallel to extract features from multi-modal inputs and can be used to predict the grasp configuration for an object.  It has been demonstrated that the use of deep residual layers helped extract better features from the input image, which were further used by the fully connected layers to output the grasp configuration. Our models improved the state-of-the-art performance on the Cornell Grasping Dataset and run at real-time speeds.

In future work, we would like to apply transfer learning concepts to use the trained model on the grasp dataset to perform grasps using a physical robot. Moreover, in an industrial setting, the detection accuracy can go even higher and can make grasp detection for pick and place related tasks robust to different shapes and sizes of parts.  


{\small
\bibliographystyle{ieeetr}
\bibliography{egbib}
}

\end{document}